\documentclass[conference]{IEEEtran}
\IEEEoverridecommandlockouts
\usepackage{amsmath,amssymb,amsfonts}
\usepackage{algorithm}
\usepackage{algorithmic}
\usepackage{graphicx}
\usepackage{textcomp}
\usepackage{xcolor}
\usepackage{booktabs}
\usepackage[square]{natbib}
\floatstyle{ruled}
\newfloat{algorithm}{tbp}{loa}
\algsetup{linenodelimiter=}

\def\BibTeX{{\rm B\kern-.05em{\sc i\kern-.025em b}\kern-.08em
		T\kern-.1667em\lower.7ex\hbox{E}\kern-.125emX}}
\begin{document}
	
	%%
	%% The "title" command has an optional parameter,
	%% allowing the author to define a "short title" to be used in page headers.
	%\title{Benign-Adversarial Joint Learning (BAJL): Towards Mitigating Robustness-Accuracy Trade-off In Deep k-Nearest Neighbors}
	
	%\title{Adversarially Enhanced Deep k-Nearest Neighbors}
	\title{Deep Adversarially-Enhanced k-Nearest Neighbors}
	
	%%
	%% The "author" command and its associated commands are used to define
	%% the authors and their affiliations.
	%% Of note is the shared affiliation of the first two authors, and the
	%% "authornote" and "authornotemark" commands
	%% used to denote shared contribution to the research.
	\author{\IEEEauthorblockN{Ren Wang}
		\IEEEauthorblockA{
			\textit{University of Michigan}\\
			renwang@umich.edu}
		\and
		\IEEEauthorblockN{Tianqi Chen}
		\IEEEauthorblockA{\textit{University of Michigan} \\
			tqch@umich.edu}
		\and
		\IEEEauthorblockN{Alfred Hero}
		\IEEEauthorblockA{\textit{University of Michigan} \\
			hero@eecs.umich.edu}}
	\maketitle
	\begin{abstract}
		Recent works have theoretically and empirically shown that deep neural networks (DNNs) have an inherent vulnerability to small perturbations. Applying the Deep k-Nearest Neighbors (DkNN) classifier, we observe a dramatically increasing robustness-accuracy trade-off as the layer goes deeper. In this work, we propose a Deep Adversarially-Enhanced k-Nearest Neighbors (DAEkNN) method which achieves higher robustness than DkNN and mitigates the robustness-accuracy trade-off in deep layers through two key elements. First, DAEkNN is based on an adversarially trained model. Second, DAEkNN makes predictions by leveraging a weighted combination of benign and adversarial training data. %trade-off between the robustness and accuracy of deep neural networks (DNNs). In this work, we first show that this trade-off enlarges dramatically when layer goes deeper. We propose a Benign-Adversarial Joint Learning (BAJL) framework that can effectively mitigate the robustness-accuracy trade-off. Build upon Deep k-Nearest Neighbor (DkNN) classifiers, BAJL makes predictions by leveraging a weighted combination of benign and adversarial training data.
		Empirically, we find that DAEkNN improves both the robustness and the robustness-accuracy trade-off on MNIST and CIFAR-10 datasets.
	\end{abstract}
	
	%% A "teaser" image appears between the author and affiliation
	%% information and the body of the document, and typically spans the
	%% page.
	% \begin{teaserfigure}
	%   \includegraphics[width=\textwidth]{sampleteaser}
	%   \caption{Seattle Mariners at Spring Training, 2010.}
	%   \Description{Enjoying the baseball game from the third-base
	%   seats. Ichiro Suzuki preparing to bat.}
	%   \label{fig:teaser}
	% \end{teaserfigure}
	
	%%
	%% This command processes the author and affiliation and title
	%% information and builds the first part of the formatted document.
	\maketitle

	% In order for DkNN++ to be useful, we need ensure that the deep representations are robust to some extent (the shift between the representations of benign examples and adversarial examples should be small enough); otherwise, the DkNN++ will suffer from a significant degradation in terms of clean accuracy
	
	\section{Introduction}
	
	Deep learning has become the backbone of many applications, such as machine translation \cite{liu2020multilingual}, image classification \cite{li2019deep}, and object detection \cite{wu2020recent}. Advanced deep neural network (DNN) models like convolutional neural network (CNN) \cite{krizhevsky2012imagenet} and recurrent neural network (RNN) \cite{graves2013speech} result in superiority over conventional models \cite{o2019deep}. As a side effect, these DNN-based applications only perform well when the input is similar to the examples in the training set, i.e., when the covariate shift of the representation is small compared with the training distribution. A large shift will result in a significant accuracy degradation \cite{ben2010theory}. More seriously, recent studies have shown that DNNs are easily fooled by adversarial examples, which are well-designed with imperceptible (to the human eyes) perturbations on benign inputs \cite{KGB16,SZS14}. The lacking of robustness and the big trade-off between the standard performance and the robustness have become bottlenecks of deep learning.
	
\begin{figure}[h]
		\centering
		\includegraphics[width=\linewidth]{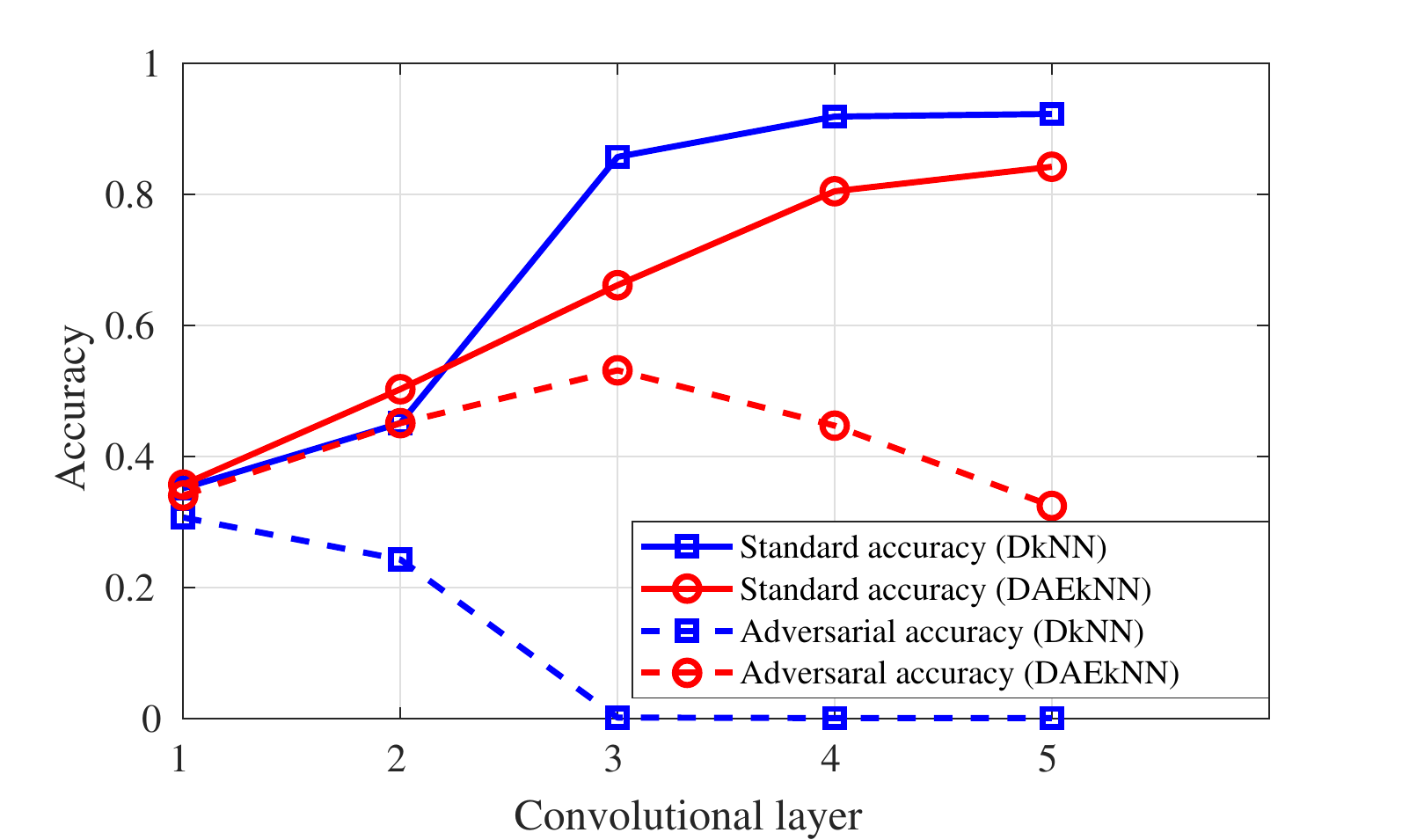}
		\caption{Deep Adversarially-Enhanced k-Nearest Neighbors (DAEkNN) can achieve higher robustness and alleviate the robustness-accuracy trade-off than applying Deep k-Nearest Neighbors (DkNN) on a model trained by standard training. All experiments are conducted on CIFAR-10. Adversarial accuracy is evaluated using $10$-step PGD attack \cite{madry2017towards} with attack power $\epsilon=8$. The blue curves show the results of DkNN. The red curves show the results of DAEkNN.}
		%\Description{trade-off.}
		\label{fig:single_layer}
	\end{figure}

	To tackle the problems, various approaches have been proposed. One category of methods is robust training, which mainly focuses on hardening model weights \cite{madry17,zhang2019theoretically,cohen2019certified} via augmenting benign examples with adversarial perturbations in the training process. This type of method is often tailored to a certain level of attack power, and is restricted to $\ell_p$ constraint perturbations. Another category of methods is the adversarial detection \cite{metzen2017detecting,feinman2017detecting,grosse2017statistical} assuming that the adversarial distribution is significantly different from the benign data distribution, which is not always true in practice \cite{carlini2017adversarial}. Strikingly, k-Nearest Neighbors (kNN)-based deep learning methods are recently demonstrated to have natural robustness defending against adversarial attacks \cite{papernot2018deep,dubey2019defense,wang2021rails} and can be leveraged to tackle other types of data quality issues beyond adversarial examples \cite{bahri2020deep}. However, the Deep k-Nearest Neighbors (DkNN) method on a standard CIFAR-10 model suffers from poor robustness and big robustness-accuracy trade-off as the layer goes deeper, as illustrated by the blue curves in Figure~\ref{fig:single_layer}. In this work, we analyze the reasons for the failure of DkNN, and propose a novel robust classifier - Deep Adversarially-Enhanced k-Nearest Neighbors (DAEkNN) that combines the kNN with the adversarial training \cite{madry2017towards} and a benign-adversarial joint prediction mechanism. We show that DAEkNN can achieve higher robustness and alleviate the robustness-accuracy trade-off compared with DkNN, as illustrated by the red curves in Figure~\ref{fig:single_layer}. 
	
	We make the following contributions in our paper.
	\begin{itemize}
		\item We show that the robustness-accuracy trade-off measured by the Deep k-Nearest Neighbors (DkNN) increases dramatically when the layer goes deeper.
		\item We propose a Deep Adversarially-Enhanced k-Nearest Neighbors (DAEkNN) method that is based on an adversarially trained \cite{madry2017towards} model and a benign-adversarial joint prediction mechanism.
		\item We demonstrate that DAEkNN enhances robustness and alleviates the robustness-accuracy trade-off. Specifically, on MNIST and CIFAR-10 datasets, DAEkNN improves DkNN robustness against PGD attack by $68.2\%$ and $50.42\%$ with $0.88\%$ and $13.34\%$ standard accuracy degradation.
	\end{itemize}

	\section{DAEKNN DESIGN}
	
	\subsection{Preliminaries}
	Throughout this paper, we evaluate different classifiers by measuring three key indices: (i) Standard Accuracy (SA) on benign test examples (ii) Adversarial Accuracy (AA) on the adversarial test examples generated by $10$-step Projected Gradient Descent (PGD) attack \cite{madry2017towards} and (iii) Harmonic Mean (HM) of AA and SA, which can be calculated by $\frac{2\text{SA}\times\text{AA}}{\text{SA}+\text{AA}}$. HM is determined by both SA and AA, reflecting the robustness-accuracy trade-off to a certain extend.
	
	\paragraph{Projected Gradient Descent (PGD) attack} PGD attack generates adversarial examples $\mathbf x_{\delta}=\mathbf x + \mathbf \delta$ to flip the prediction via maximizing the cross-entropy loss $L$. The generator implements gradient ascent with projection, with the $i$th iterations defined as 
	\begin{align}\label{knn_att_grad}
	\mathbf x_{\delta}^{i+1} \leftarrow \prod_{\mathcal{C}(\mathbf x_{\delta},\epsilon)} \big[ \mathbf x_{\delta}^i + \kappa {\rm{sign}}\big(\nabla_{\mathbf x_{\delta}}  L(f_\theta, \mathbf x_{\delta}^i, y)\big)\big] 
	\end{align}
	where $f_\theta$ is the mapping from DNN input to output. %$L(f_\theta, \mathbf x_{\delta}^i, y)$ is the cross-entropy loss. 
	$\kappa$ is the gradient step size. $\prod$ denotes a projection operator, and $\mathcal{C}(\mathbf x_{\delta},\epsilon) = \{\mathbf x_{\delta}| {\| \mathbf \delta\|_p} \le \epsilon, \mathbf x_{\delta}  \in [v_{\text{lb}},v_{\text{ub}}]^d \}$ represents the projection set. Here $\epsilon$ is a positive constant, and we use $\|\cdot\|_p=\|\cdot\|_\infty$ in this paper. If not otherwise specified, we use the attack power $\epsilon=8/80$ for CIFAR-10/MNIST. $d$ is the input dimension. $[v_{\text{lb}},v_{\text{ub}}]$ is the constraint on the data domain. In this paper, we consider $\mathbf x_{\delta}$ as an image vector, thereby $\mathbf x_{\delta} \in [0,255]^d$.
	
	\paragraph{Deep k-Nearest Neighbors (DkNN)} Different from DNN classifiers, DkNN \cite{papernot2018deep} combines kNN classifiers that are embedded into selected layers of a DNN, and makes predictions via majority vote among all kNNs. For any test input $\mathbf x$, DkNN classifies it to
	\begin{equation}\label{eq: dknn}
	\begin{aligned}
	y =   \arg\max_c \sum_{l \in \mathcal{L}} p_l^c(\mathbf x; D_{\text{tr}}^b),  ~ c \in [C],
	\end{aligned}
	\end{equation}
	where $\mathcal{L}$ is the set of the selected layers and $l \in \mathcal{L}$ is the $l$-th layer of a DNN. $f_{\theta_l}$ denotes the mapping from input to layer $l$. Here $p_l^c(\mathbf x)$ denotes the confidence score assigned to class $c$ predicted by the corresponding kNN in layer $l$. We use $[C]$ to denote the set $\{1,2,\cdots,C\}$, where $C$ is the number of classes. $D_{\text{tr}}^b=(X_{tr}^b, Y_{tr}^b)$ is the selected training dataset containing all benign examples $X_{tr}^b$ together with their labels $Y_{tr}^b$, and $D_{\text{tr}}^b(j)=(X_{tr}^b(j), Y_{tr}^b(j))=(\mathbf x_j, y_j)$ represents the $j$th example in $D_{\text{tr}}^b$. Let $K_c^l$ be the number of data points belonging to class $c$ among the $K$ nearest neighbors found via measuring $\text{dist}\big(f_{\theta_l}(\mathbf x), f_{\theta_l}(X_{tr}^b(j))\big), \forall j \in |X_{tr}^b|$, where $|\cdot|$ means the cardinality of a set. We use $\ell_2$ distance for $\text{dist}(\cdot, \cdot)$ by default. The confidence score can be calculated by $p_l^c(\mathbf x; D_{\text{tr}}^b) = \frac{K_c^l}{K}$.
	
	By default, DkNN is applied on a model trained by standard training. From Figure~\ref{fig:single_layer}, one can see that the adversarial accuracy obtained by DkNN drops dramatically as the layer goes deeper. As the standard accuracy obtained by DkNN monotonically increases, the robustness-accuracy trade-off also rises. To improve both the robustness and robustness-accuracy trade-off, we introduce a Deep Adversarially-Enhanced k-Nearest Neighbors (DAEkNN) method containing two key components: (i) A DNN model obtained by adversarial training (ii) A joint prediction mechanism leveraging benign and adversarial training data. We introduce each of them in detail below.

	\begin{figure}[h]
		\centering
		\includegraphics[width=\linewidth]{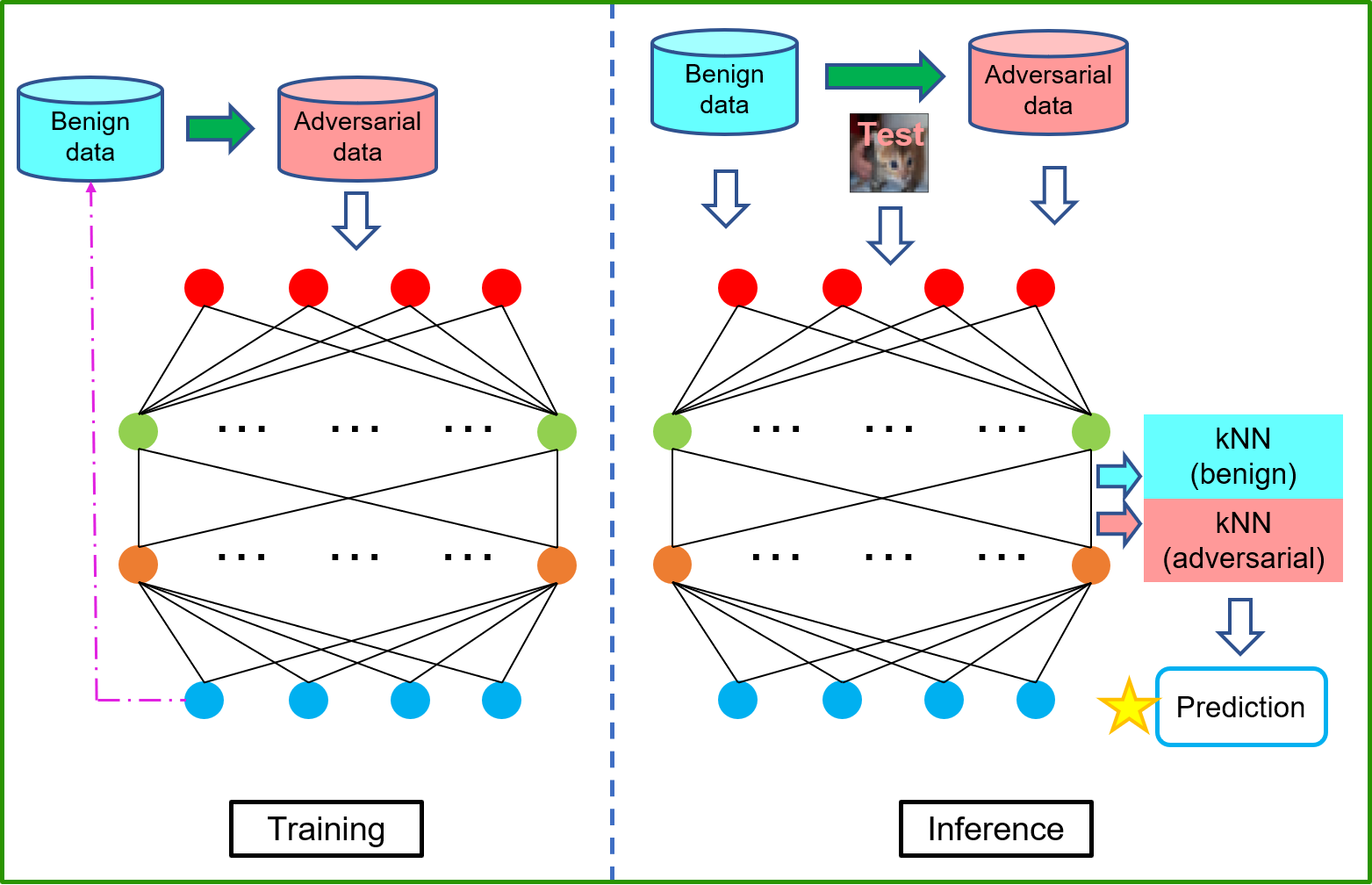}
		\caption{The framework of DAEkNN: DAEkNN is based on an adversarially trained deep neural network (DNN) model and a benign-adversarial joint prediction mechanism. In the training phase, a DNN is trained with adversarial data augmentation \cite{madry2017towards}. In the inference phase, the prediction is made by a joint classifier applied on deep layers of the adversarially trained DNN and combined the k-nearest neighbors (kNN) classifiers based on benign training data and its adversarial counterpart.}
		%\Description{framework.}
		\label{fig:framework}
	\end{figure}

	\subsection{Improving Robustness Through Adversarially Trained Model}
	From Figure~\ref{fig:single_layer}, one can see that the adversarial accuracy of DkNN drops dramatically as the layer goes deeper, indicating the features learned by DNN lack robustness. In order to robustify k-nearest neighbor classifiers in deep layers, we need the features extracted from those layers to be robust beforehand. Recent works have demonstrated that models trained by standard training tend to learn non-robust features, while models trained by adversarial training \cite{madry2017towards} are able to extract more robust features \cite{engstrom2019adversarial,wang2021fast,wang2020practical}.
	
	As shown in Figure~\ref{fig:framework}, the adversarial training is performed by updating model weights with adversarial examples generated from benign data in each epoch, and can be characterized by the following optimization problem.
	
	\begin{align}\label{eq: AT}
	\begin{array}{ll}
	\arg\min_\theta \frac{1}{|S|} \sum_{(\mathbf x, y) \in S}\max_{\|\mathbf \delta\|_\infty \le \epsilon_{\text{tr}}} L(f_\theta, \mathbf x + \mathbf \delta, y),
	\end{array}
	\end{align}
	where $\epsilon_{\text{tr}}$ is the attack power of the adversarial generator in the training process, and in this paper we fix it to $\epsilon_{\text{tr}}=4$ and $\epsilon_{\text{tr}}=60$ for CIFAR-10 and MNIST, respectively. $S$ denotes a subset of $D_{\text{tr}}^b$. We remark that there are different robust training methods \cite{madry2017towards,zhang2019theoretically} that can be leveraged to harden DNNs, and we only focus on the adversarial training method \cite{madry2017towards}.
	
	\subsection{Improving Robustness Trough Adversarial Training Data}\label{sec: adv_tr}
	Note that adversarial perturbations are intended to enlarging the gap between the adversarial test distribution and the distribution of the original benign class. Using benign training distribution as the baseline, the covariate shift of the adversarial test distribution is larger than the covariate shift of the benign test distribution. For this reason, when all k-nearest neighbors are found from benign training data, the kNN classifiers are not able to correctly capture the information of the adversarial distribution. Spurred by that, we ask: Is it possible to better handle the covariate shift as well as obtaining more robust classifiers by changing the baseline (benign) distribution to a benign-adversarial combined distribution (in other words, by adding adversarial examples into the kNN searching space)? 
	
	Instead of using \eqref{eq: dknn}, we propose a benign-adversarial joint prediction mechanism that robustifies classifiers through incorporating the adversarial training data, as illustrated in the inference stage in Figure~\ref{fig:framework}. Mathematically, the benign-adversarial joint prediction is expressed by the following formula.
	
	\begin{align}\label{eq: daeknn}
	\begin{array}{ll}
	y =   \arg\max_c \sum_{l \in \mathcal{L}} \big[ \omega_a^l p_{l}^c(\mathbf x; D_{\text{tr}}^a) + \omega_b^l p_{l}^c(\mathbf x; D_{\text{tr}}^b) \big],   c \in [C],
	\end{array}
	\end{align}
	where $D_{\text{tr}}^a$ denotes the adversarial training data obtained by applying PGD attack on $D_{\text{tr}}^b$.
	\begin{align}\label{eq: adv_td}
	\begin{array}{ll}
	D_{\text{tr}}^a =   \text{PGD}(\theta, \epsilon_r, D_{\text{tr}}^b),
	\end{array}
	\end{align}
	Here $D_{\text{tr}}^a$ serves as the hardening dataset. We therefore call $\epsilon_r$ the hardening strength. We remark that $\epsilon_r$ is not as larger as better. For DAEkNN to be useful, we need to ensure that the features extracted by deep layers contain useful class information. For example, a model trained by standard training cannot tolerate large perturbations, and therefore using a large $\epsilon_r$ will hurt the performance. In contrast, an adversarially trained model can extract class-relevant features from both benign data and perturbed data. In this sense, an adversarially trained model results in narrower gap between the representations of benign examples and the adversarial examples. $\omega_a^l$ and $\omega_b^l$ are weights of the two classifiers. Note that \eqref{eq: daeknn} is not a simple mixed up of the benign and adversarial training data. Instead, the new classifier in a single layer is a weighted combination of two classifiers. The weights $\omega_a^l$, $\omega_b^l$ are calculated based on the relative distance of the kNN found by the two classifiers.
	
	\begin{align}\label{eq: weight}
	\begin{array}{ll}
	\omega_a^l = \frac{e^{\text{dist}\big(f_{\theta_l}(\mathbf x), f_{\theta_l}(X_l^b)\big)}}{e^{\text{dist}\big(f_{\theta_l}(\mathbf x), f_{\theta_l}(X_l^b)\big)} + e^{\text{dist}\big(f_{\theta_l}(\mathbf x), f_{\theta_l}(X_l^a)\big)}}, \\ \omega_b^l = \frac{e^{\text{dist}\big(f_{\theta_l}(\mathbf x), f_{\theta_l}(X_l^a)\big)}}{e^{\text{dist}\big(f_{\theta_l}(\mathbf x), f_{\theta_l}(X_l^a)\big)} + e^{\text{dist}\big(f_{\theta_l}(\mathbf x), f_{\theta_l}(X_l^b)\big)}}
	\end{array}
	\end{align}
	where $X_l^b$ denotes the kNN of $\mathbf x$ found from $D_{\text{tr}}^b$ on layer $l$, and similarly, $X_l^a$ denotes the kNN of $\mathbf x$ found from $D_{\text{tr}}^a$ on layer $l$. We use the softmax version of the distance-based weights. One can see that the weight is large when the corresponding distance is small, and $\omega_a^l+\omega_b^l=1$.
	
	\eqref{eq: daeknn} refines the decision boundaries of \eqref{eq: dknn} by leveraging the combined and weighted classifiers. The implementation details can be viewed in Algorithm \ref{alg: joint-classifier}. We also refer readers to Section \ref{sec:experiment} for the ablation study and comparisons.

	\begin{algorithm}[h]
		\caption{Deep Adversarially-Enhanced k-Nearest Neighbors (DAEkNN)}
		\label{alg: joint-classifier}
		\begin{algorithmic}
			\REQUIRE Model $\theta$ trained by adversarial training with $\epsilon_{\text{tr}}$; the selected layers $\mathcal{L}$; hardening strength $\epsilon_r$; test input $\mathbf x^{\prime}$; number of nearest neighbors $K$.
			\STATE Generate $D_{\text{tr}}^a$ from $D_{\text{tr}}^b$ through \eqref{eq: adv_td}.
			\FOR{All $l \in \mathcal{L}$}
			\STATE Find $K$ nearest neighbors $X_l^b$ and $X_l^a$ from $D_{\text{tr}}^b$ and $D_{\text{tr}}^a$, respectively.
			\STATE Count the number of data belonging to class $c$ ($K_c^l$) among each of $X_l^b$ and $X_l^a$.
			\STATE Calculate $p_{l}^c(\mathbf x^{\prime}; D_{\text{tr}}^b)$ and $p_{l}^c(\mathbf x^{\prime}; D_{\text{tr}}^a)$ using $\frac{K_c^l}{K}$ obtained from $X_l^b$ and $X_l^b$.
			\STATE Calculate $\omega_a^l$, $\omega_b^l$ through \eqref{eq: weight}.
			\ENDFOR
			\STATE Obtain the final prediction by \eqref{eq: daeknn}.
			
			\STATE {\bfseries Output:} $y$.
		\end{algorithmic}
	\end{algorithm}
	
	\section{Experimental Results}\label{sec:experiment}
	We compare DAEkNN with standard Convolutional Neural Network (CNN), Adversarially Trained \cite{madry2017towards} Convolutional Neural Network (AT-CNN), and Deep k-Nearest Neighbors (DkNN) \cite{papernot2018deep} on MNIST and CIFAR-10 datasets. We implement a three-convolutional-layer architecture for MNIST and the VGG16 architecture for CIFAR-10. VGG16 has five convolutional layer blocks and we call the $i$th block as convolutional layer $i$. By default, we set the attack power $\epsilon_r=8/80$ and the hardening strength $\epsilon_r=8/60$ for CIFAR-10/MNIST. The performances are measured by Standard Accuracy (SA) evaluated using benign examples, Adversarial Accuracy (AA) evaluated using adversarial examples, and the Harmonic Mean (HM) of SA and AA. %We generate adversarial examples via $10$-step PGD attack \cite{madry2017towards} with $\epsilon=8$.
	
	Table~\ref{tab:overall_performance} shows the comparisons of different methods. DAEkNN improves DkNN / AT-CNN / CNN robustness against PGD attack by $68.2\%/5.98\%/68.26\%$ and $50.42\%/18.86\%/50.42\%$ with only $0.88\%/0.5\%/1.16\%$ and $13.34\%/7.12\%/13.8\%$ standard accuracy degradation on MNIST and CIFAR-10, respectively. We also compare DAEkNN with DkNN on different layers of VGG16, as shown in Figure~\ref{fig:single_layer}. One can see that DAEkNN improves DkNN on both the robustness and robustness-accuracy trade-off.
	
	\begin{table}
		\caption{Deep Adversarially-Enhanced k-Nearest Neighbors (DAEkNN) outperforms Convolutional Neural Network (CNN), Adversarially Trained \cite{madry2017towards} Convolutional Neural Network (AT-CNN), and Deep k-Nearest Neighbors (DkNN) \cite{papernot2018deep} in terms of Adversarial Accuracy (AA) without appreciable loss of Standard Accuracy (SA) on MNIST and CIFAR-10. Both DAEkNN and DkNN are applied on convolutional layer $2, 3$ for MNIST and $3, 4$ for CIFAR-10.}
		\centering
		\label{tab:overall_performance}
		\begin{tabular}{lccccc}
			\toprule
			& & CNN & AT-CNN & DkNN & \textbf{DAEkNN}\\
			\midrule
			MNIST & SA & \textbf{99.06}\% & 98.4\% & 98.78\% & 97.9\% \\
			& AA & 0\% & 62.28\% & 0.06\% & \textbf{68.26}\% \\
			& HM & 0\% & 76.28\% & 0.12\% & \textbf{80.44}\% \\
			\midrule
			CIFAR-10 & SA & \textbf{92.24\%} & 85.56\% & 91.78\% & 78.44\% \\
			& AA & 0.06\% & 31.62\% & 0.06\% & \textbf{50.48\%} \\
			& HM & 0.12\% & 46.18\% & 0.12\% & \textbf{61.43\%} \\
			\bottomrule
		\end{tabular}
	\end{table}
	
	We also consider leave-one-out versions of DAEkNN, which can be viewed as an ablation study. The two variants of DAEkNN are DAEkNN without adversarial training component (DAEkNN-WAT) and DAEkNN without adversarial training data (DAEkNN-WAD). As explained in Section \ref{sec: adv_tr}, the hardening strength $\epsilon_r$ needs to be set small for a model trained by standard training. We therefore set $\epsilon_r=2$ for DAEkNN-WAT. One can see from Table~\ref{tab:overall_harmonic_mean} that even DAEkNN-WAT can significantly improve the robustness of DkNN. With adversarial training data, DAEkNN improves the robustness of DAEkNN-WAD by $3.82\%$.  
	
	\begin{table}
		\caption{Comparisons among Deep k-Nearest Neighbors (DkNN) \cite{papernot2018deep}, Deep Adversarially-Enhanced k-Nearest Neighbors (DAEkNN), DAEkNN without adversarial training component (DAEkNN-WAT), and DAEkNN without adversarial training data (DAEkNN-WAD). Performances are measured by Adversarial Accuracy (AA), Standard Accuracy (SA), and the Harmonic Mean (HM) of AA and SA on CIFAR-10. All methods are applied on convolutional layer $3, 4$.}
		\label{tab:overall_harmonic_mean}
		\centering
		\begin{tabular}{lcccc}
			\toprule
			& DkNN & \begin{tabular}[c]{@{}c@{}} \textbf{DAEkNN} \\ \textbf{-WAT}   \end{tabular} & \begin{tabular}[c]{@{}c@{}} \textbf{DAEkNN} \\ \textbf{-WAD}   \end{tabular} & \textbf{DAEkNN} \\
			\midrule
			SA & \textbf{91.78\%} & 85.96\% & 79.9\% & 78.44\% \\ 
			AA & 0.06\% & 13.1\% & 46.66\% & \textbf{50.48\%} \\
			HM & 0.12\% & 22.74\% & 58.91\% & \textbf{61.43\%} \\
			\bottomrule
		\end{tabular}
	\end{table}
	
\begin{figure}[h]
		\centering
		\includegraphics[width=\linewidth]{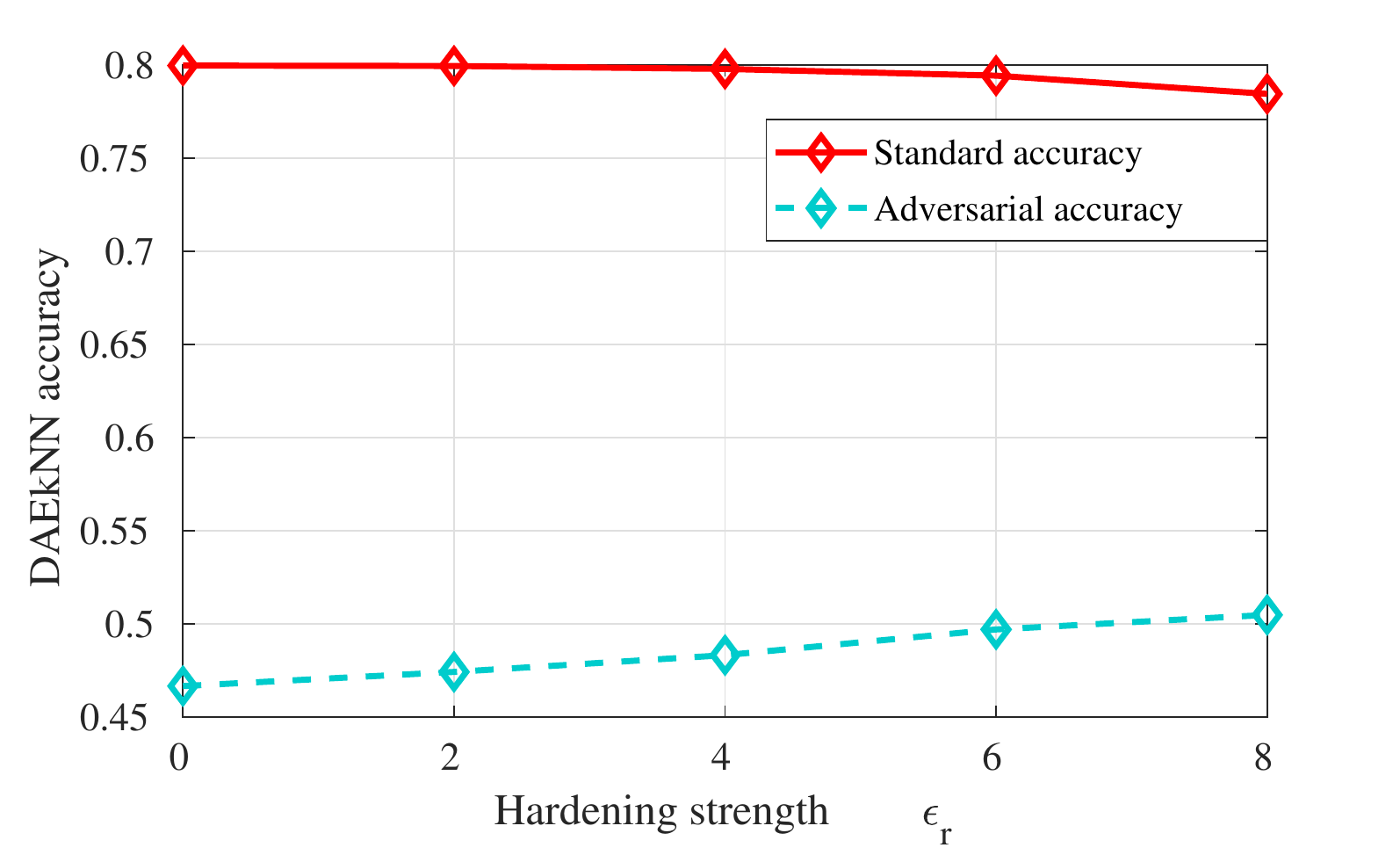}
		\caption{Increasing the hardening strength $\epsilon_r$ within a certain range ($\le \epsilon$ in this test) can improve the robustness of Deep Adversarially-Enhanced k-Nearest Neighbors (DAEkNN) without appreciable loss of standard accuracy. All experiments are conducted on CIFAR-10. Adversarial accuracy is evaluated using a $10$-step PGD attack \cite{madry2017towards} with attack power $\epsilon=8$.}
		%\Description{hardening strength.}
		\label{fig:epsilon}
	\end{figure}

	We then study how the hardening strength $\epsilon_r$ affects the performance of DAEkNN. As shown in Figure~\ref{fig:epsilon}, increasing $\epsilon_r$ from $0 - 8$ improves the robustness without appreciable loss of standard accuracy.
	
	\section{CONCLUSION}
	In this work, we propose a new deep neural network (DNN)-based classifier, called Deep Adversarially-Enhanced k-Nearest Neighbors (DAEkNN), which enjoys high robustness and can mitigate the robustness-accuracy trade-off. We first study the  performance of single-layer Deep k-Nearest neighbors (DkNN) on each layer of a DNN. The results indicate that DkNN has poor robustness and big robustness-accuracy trade-offs. The proposed DAEkNN improves robustness through two key adversarial components. DAEkNN extracts features from an adversarially trained model that can provide more robust features. DAEkNN utilizes both the benign training data and adversarial training data in a benign-adversarial joint prediction classifier. The experiments on MNIST and CIFAR-10 demonstrate the effectiveness of DAEkNN.
	
	In future work, we will (i) consider more attack types; (ii) explore the combinations of adversarial training data generated by different hardening strength and different attack types, which could potentially improve the robustness under stronger and unforeseen attacks; (iii) provide a theoretical analysis of the DAEkNN.

	%%
	%% The next two lines define the bibliography style to be used, and
	%% the bibliography file.
	%\bibliographystyle{ACM-Reference-Format}

	\bibliography{reference,refs_adv}

\end{document}